\documentclass[sigconf]{acmart}

\usepackage{booktabs} 

\usepackage{algorithm2e}

\acmArticle{4}
\acmPrice{15.00}

\editor{Jennifer B. Sartor}
\editor{Theo D'Hondt}
\editor{Wolfgang De Meuter}

\begin{document}
\title{Video Based Contextual Q\&A}

\author{Akash Ganesan}
\affiliation{%
  }
\email{akaberto@umich.edu}

\author{Divyansh Pal}
\affiliation{%
  }
\email{divpal@umich.edu}

\author{Karthik Muthuraman}
\affiliation{%
  }
\email{mkarthik@umich.edu}

\author{Shubham Dash}
\affiliation{%
  }
\email{shubhamd@umich.edu}


%
%

\ccsdesc[500]{Computing methodologies~Object detection}
\ccsdesc[500]{Computing methodologies~Natural language generation}
\ccsdesc[500]{Computing methodologies~Scene understanding}
\ccsdesc[500]{Computing methodologies~Visual content-based indexing and retrieval}

\maketitle

\section{Abstract}

The primary aim of this project is to build
a contextual Question-Answering model for videos. The
current methodologies provide a robust model for image based
Question-Answering, but we are trying to generalize
this approach to be videos. We propose a graphical representation of video which is able
to handle several types of queries across the whole video. For example,
if a frame has an image of a man and a cat sitting,
it should be able to handle queries like, where is the cat
sitting with respect to the man? or ,what is the man holding
in his hand?. It should be able to answer queries relating to
temporal relationships also.

\section{Introduction}
The big data explosion, combined with the availability of a large
amount of labeled data and better hardware has spurned large
scale deep learning.  At intersection of natural language
processing and computer vision is understanding of context in
images and videos.  While extensive work has been done with
understanding images, not much has been done on videos.  The
recent trend involves use of LSTMs for understanding the whole context
of a given frame in a video and stitching them together.  While this
preserves the temporal information of the whole video, it doesn't let us query  videos based on the contextual information present.

In this project, we extend this work into graph domain by encoding the context of these videos as a graph using research work done by the computer vision community, which can handle more contextual information than any other representation.

Based on a graph representation of a video which preserves context, we look into a few other applications like extracting key-frames based on high context change observed between frames using graph similarity measures. Also we look into video context similarity at a high level to propose how the knowledge of contextually similar videos can be used to cluster similar videos and can lead to recommending videos on platforms like YouTube, which currently recommend videos based on human annotated tags for a video.

The pipeline for this project is as follows:
\begin{enumerate}
\item 
Frame Extraction - Videos are converted to image frames using frame extraction algorithm.

\item
Object Detection and Captioning- Main objects is detected in each frame using R-CNN technique. Dense Captioning algorithm is used to caption relationship of these objects.

\item
Scene Graph Generation - Each caption is converted to a scene graph using standard NLTK python library.

\item
Graph Aggregation - The scene graphs are aggregated to get a scene graph for each frame. These are then appended with a time stamp. Finally all the scene graphs are aggregated to give an aggregate graph for the whole video. Various similarity measures are used to compare scene graphs and aggregated video graphs.
\end{enumerate}

\section{Data}
For our experiments, videos of varying lengths and context were chosen from the YouTube 8M dataset. 5 videos were used for testing the querying model.

\vspace{1mm}

\noindent The videos in consideration include instructional videos where a particular recipe is followed, a music video with lots of activity and scene changes, and a video where a girl trying out food in a restaurant.

\vspace{1mm}

\noindent The average statistics for the videos are:

\vspace{1mm}

\begin{itemize}
\item The average video length around 8 minutes (~500 seconds)
\item With uniform sampling of about 0.5 seconds for each samples, number of frames extracted for each video is about 1000 frames.
\item The aggregate scene graph contains on an average around 300 nodes and about 800 edges.

\end{itemize}
For our current experiments as well as module testing, we have taken common stock images and videos which encompass a wide variety of variations in terms of context and relationship among the objects as well as a rich set of attributes for many objects in question.

\section{Methodology}
In this section, we go over in detail all the sequential blocks used in our pipeline and how they are linked. On a high level, we have the following tasks : frame extraction, object detection and captioning, scene graph generation, scene graph alignment and aggregation.
We also find key-frames from the frame scene graphs using graph similarity measures.
We will further explore converting an incoming query into a query scene-graph and graph based search methods to answer the question. 

\subsection{Frame Extraction}


We use a few of the frame extraction techniques in our experiments. Firstly we use uniform sampling (every 0.5 seconds) for prototyping purposes. Later for the final experiments we also make use of the mean absolute difference between pixel values of adjacent frames to detect a change. 

\subsection{Object Detection and Captioning}

Once we have the frames, we extract from each frame relevant captions and bounding box information. This is done using the DenseCap\cite{DenseCap} model. The image is first passed through a Regional-Convolutional Neural Network which detects key objects and returns bounding boxes and confidence scores. The object labels and actions are then passed through a Recurrent Neural Network which generates a one-sentence caption for a particular object detected.

We repeat this for each of the key frames and obtain multiple captions per frame and use that as an input for the next block in our pipeline.

\subsection{Scene Graph Generation}
This module converts each caption from Object Detection and Captioning module and convert it into a scene graph.
On the representation side, two parsers have been discussed in the
paper \cite{s2}, namely, rule-based parsers and classifier-based parsers
to generate scene graphs from text. Both work on an alternative
linguistic representation known as semantic graph. While Rule $-$
based parsers work on a combination of nine dependency patterns based on Semgrex \cite{DBLP:conf/depling/Tamburini17} expressions, which matches text to a dependency tree based on Part-of-Speech tagging, these include:
\begin{itemize}
\item Adjectival modifiers
\item Copular constructions
\item Prepositional phrases
\item Possessive constructions
\item Passive constructions
\item Clausal modifiers of nouns
\end{itemize}

Classifier-based parsers learn the relationship between individual
pair of words (x1, x2) which are classified on object features, lexicalized
features and syntactic features using an L-2 regularized maximum
entropy classifier for classification. The classifier classifies words into `subject', `object' and `predicate' which form the nodes and edges of our scene graph.
The code implementation we use follows the classifier based parser method which provides better accuracy. 

\begin{figure}[h!]
\centering
\includegraphics[width = 1\linewidth]{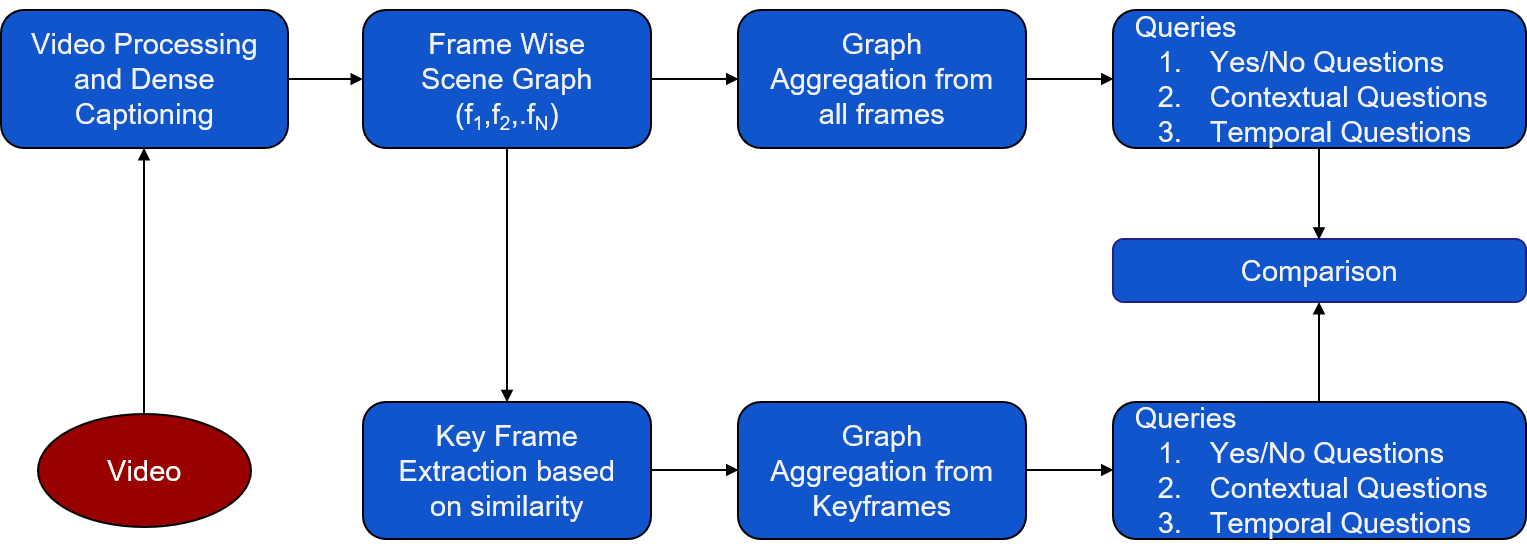}
\caption{Proposed Methodology}
\label{fig:pipeline}
\end{figure}

\subsection{Similarity measures}

Two methods have been primarily looked at for the purpose of measuring similarity between graphs. Similarities being considered in the project have a two fold use. The first is similarity between graphs which is used as a measure to compare videos based on contextual content which can be further used to cluster contextually similar videos together. The second reason of applying similarity measures to frame by frame representations of scene graphs, is to extract key-frames which allows for a much more compact and reasonable contextual representation of the video. As an added application, the current methods for extracting key-frames are pixel based, so this can act as a more context encoding alternative for key-frame extraction.

\noindent The similarity measures are as follows:

\begin{itemize}
\item \textbf{Spectral score:} The Frobenius norm of the difference of the adjacency matrix representation of the graphs has been used as a similarity score. Such similarity score would be able to encapsulate whether or not a subgraph has a "similar" neighborhood and the Frobenius norm will penalize any differences or changes in the neighborhood of a particular node. For example a subgraph having nodes "tall $-$ man $-$ wearing $-$ black $-$ T-shirt" would have a different adjacency matrix to "thin $-$ man $-$ feeding $-$ dog".

\vspace{2mm}
The similarity score is defined as:

\textit {sim\_score} $=$ 1 $-$ $\frac{||{A - B}||_F}{n}$

\vspace{2mm}
\noindent This similarity score will give a similarity score between 0 and 1, 1 indicating perfect similarity between two graphs $G_1$ and $G_2$, of whom A and B are adjacency matrix as $\frac{||{A - B}||_F}{n}$ will be a zero quantity. 
Similarly, 0 indicates perfect dissimilarity between the graphs as $\frac{||{A - B}||_F}{n}$ will be one.

The space and time complexity of this method of similarity comes out to be $O(n^2)$  

\item \textbf{Maximum Common Subgraph:} This method finds the largest subgraph of $g_{1}$ and $g_{2}$. It is a method to check how similar are the graphs structurally. A common subgraph of $g_1$ and $g_2$ is a graph $g$ such that there exist subgraph isomorphisms from $g$ to $g_1$ and from $g$ to $g_2$. A graph $g = mcs(g_1,g_2)$ is maximum common subgraph of $g_1$ and $g_2$, if there exists no other common subgraph of $g_1$ and $g_2$ that has more nodes than $g$.\\
The similarity score is defined as follows:\\
\textit {sim\_score} $= \frac{|mcs(g_{1},g_{2})|}{|g_{1}| + |g_{2}| - |mcs(g_{1},g_{2})|}$\\
where,\\
$g_1$, $g_2$ are two graphs.\\
$mcs (g_1,g_2)$ is the maximum common sub graph of two graphs where mcs is the largest graph (by some measure involving the number of nodes and edges )contained in both subject graphs.\\
|$g_1$|, ${|g_2|}$ are cardinality of the graph $g_1$ and $g_2$.

The space and time complexity of this algorithm is roughly estimated to be $O(2^n)$

\end{itemize}

\subsection{Graph Formulation}
The graph formulated for the experiment can be visualized as a tripartite graph.
The graph consists of three types of nodes describing:
\begin{itemize}
\item Subjects like man, dog, cat etc.
\item Attributes like tall, fat, brown etc.
\item Relationships like feeding, throwing etc.
\end{itemize}

A sample graph formulation is shown below where nodes in red are subjects, the nodes in yellow are relationships and nodes in green are attributes.
It can be observed that a common path in the graph is subject $-$ relationship $-$ subject, with attributes as leaf nodes, meaning no outgoing edges.

\vspace*{4mm}

\includegraphics[width = 1\linewidth]{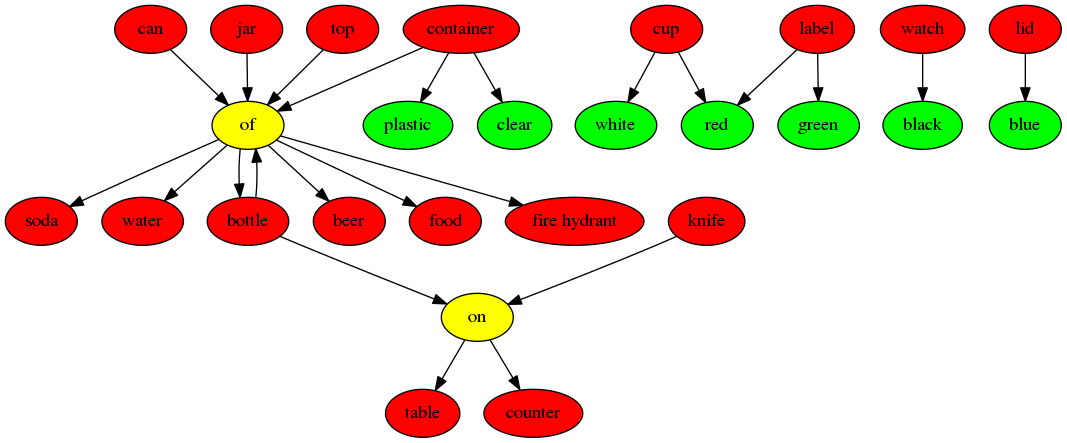}

\vspace*{2mm}
\noindent The four primary graph that are formulated to handle queries of different types are listed as follows:

\begin{itemize}
\item \textbf{Individual frames to scene graph}: This representation is used to handle temporal questions and extract key frames in a video based on frame similarity basis.
\item \textbf{A bag-of-nodes based scene graph aggregation}: In order to form a cumulative graph representation for the entire video. This representation is useful for calculating video similarity across different videos and cluster similar videos based on content.
\item \textbf{Scene Graph aggregation based on key-frames}: Building an aggregate scene graph based on key-frames helps answer queries on the major recurring theme of the video or the plot of the video. 
\item \textbf{Node similarity based scene graph aggregation}: This graph formulation allows for a more compact graph representation for the video along with preserving the uniqueness of multiple instances of a class, based on the attributes. For instance this representation will help uniquely identify to different men appearing in different scenes in a video.
\end{itemize}

The formulations are discussed in more detail below

\subsubsection{\textbf{Conversion of individual frames to scene graphs:}}

As frames sampled in a video describe scenes for those timestamps, a scene graph representation of these video frames would be useful for answering queries which involve temporal information. Queries like "what did the man eat before going home" can be answered using this approach as traversing over time slices that occur before subgraph man $-$ driving and will allow to localize all the actions performed by man before the time slice. 

\subsubsection{\textbf{Bag-of-nodes aggregation of scene graphs:}}
This representation allows for building a high-level contextual representation of the video. A salient feature of this representation would be that does not distinguish between different instances of the same class. Man occurring in different scenes are assumed to the same and are not differentiated based attributes. The intuition behind this kind of formulation is similar to measuring sentence similarity using a bag words of approach, as a high-level content similarity in video is measured across different type of classes occurring in the scene as opposed to unique occurrences of the same class. The video similarity is measured using a spectral decomposition method and the maximum subgraph matching method, as discussed earlier.

\subsubsection{\textbf{Scene Graph aggregation based on Node similarity:}}

In this section we propose a novel approach to build an aggregated `scene' graph which captures both temporal and relational edge attributes.
More formally at the end of the video, we will have a heterogeneous graph $G(E,V,T)$ where each Node (V) represents an instance of an object (like man, cat, dog), each edge (E) represents the relationship between the objects and is appended with a time stamp and each node has attributes(T) which captures the type of that object. Additionally store this tensor representation of the graph for each time-stamp for easy retrieval/querying. 

The graph we are building will be a time-evolving heterogeneous semantic graph. Many existing node similarity measures have been explored for the this project and have some inherent shortcomings. For eg: FINAL (Fast attributed network alignment) deals with node attributes and edge attributes, but deal primarily with homogeneous matrices. Exploring similarity measures for heterogeneous graphs like metapath2vec[2] also lead to a dead-end. This was due to the huge variations in relationships that are captured in the graph, making it extremely difficult to define a metapath which will be the right fit for calculating node similarities for the entire graph. For eg: Consider two similar nodes say man (man\_1 and man\_2), both of them are connected to a dog, but they form different relations with the dog. In one case the man might be feeding the dog, and in the other case, man might be throwing a ball. So to conclude any metapath describing a object1-object2-object3 path (for example man-dog-man) will fail here. Moreover we do not have a small class of distinct objects like in the academic citation network mentioned in metpath2vec. This effectively rules out metapath2vec as a viable similarity technique in this work.

We thus propose the following node similarity measure between two nodes of the same class (say woman).
Let
\begin{equation*}
\tau(x) = cardinality \; of \; the \; set \; x
\end{equation*}
Let attr\_u and attr\_v be the set of attributes of u and v respectively. 
Then,
\begin{equation*}
NodeSim(u,v) = \frac{\tau(attr\_u \cap attr\_v)}{\tau(attr\_u \cup attr\_v)}
\end{equation*}
Note that the above similarity measure is a modification on Jaccard's score.

Now we also observe that the above function can be used to measure similarity of two nodes of same class (example woman\_1 and woman\_2 of same class woman). However, this is for the two women belonging to a different frame. For the same frame when we have captions like "Woman is tall", "Woman is driving" etc we may want to know which woman is being referred to within a single frame. 
For this distinction, we make use of the bounding box information that we have obtained during the dense captioning. Similar to the above NodeSim formula, we can use the Intersection-over-Union (IoU) metric to evaluate similarity between objects in the same frame. 

Now that we have defined our node similarity methods, we describe the scene graph aggregation method below. 
Let $G(E,V,T,time)$ denote the aggregated scene graph so far. Let $G_1(E_1,V_1,T_1)$ denote an incoming scene graph.

\begin{algorithm}[h]
\SetAlgoLined
\KwResult{Aggregated Scene Graph G}
 \For{every $v_1$ in $G_1$}{
  \For{past m time instances}{
   \For{every $v \epsilon G(E,V,T,m)$ and v belongs to same object class of $v_1$}{
  \eIf{$NodeSim(v,v_1) < threshold$}{
   Create new node $v_1$ in G(E,V,T,m)\;
   Create all the edges corresponding to $v_1$\;
   }
   {
   Map edges of $v_1$ to v;
  }
  }
	}
    }
 \caption{Scene Graph Aggregation}
\end{algorithm}

Note that if $v_1$ is the first instance of an object type, it is added as a new node by default. The above algorithm will only decide when a repeated instance of an object type (like woman\_1 and woman\_2) should be considered as a new node.

In the above algorithm, m and threshold are hyper-parameters which can be tweaked to obtain acceptable results. This way we build our aggregate scene graph.

Note that in the beginning, we pre-process the scene graph for every frame and obtain one dictionary(hash table) for each containing unique attributes and unique relationships. This will enable us to one-hot encode any incoming attribute or relationship uniquely. This is what the third dimension (T) of the tensor contains. 

\subsection{Query Formulation}
For this project, we are looking into three types of question that will be used to test the graph formulation. Customized functions have been made for each type of query. These are:

\begin{itemize}
\item True/False questions - The purpose of this type of question is to check if a subgraph of question exists in the graph or not. The input to function in these sort of questions is two objects and an edge. The output is True or False based if the subgraph formed by object - edge - object exists in aggregated graph or not.
\item Contextual questions - These answer questions like where, who and what? The input to such type of questions are one object and edge. The answer in such question are all objects which complete object- edge - object subgraph. The answer is usually a list of all such objects. For example, if input is man - wear, then the answer is [suit, tie, hat] as all three subgraphs exist in the aggregated graph.
\item Temporal questions - These answer questions like did something happen before or after something else. The input for such questions is two sequence of events. The output is true is sequence\_{1} happens before sequence\_{2}. For example if input is main - eat - pizza - man - play - dog, then the output is True if man ate pizza before man played with dog and false if either sequence did not happened or the sequence happened other way round. 
\end{itemize}

\noindent The question is converted to object - edge - object graph by using NLTK python library and spacy package. It uses Part-of-Speech tagging to find nouns and verbs in a sentence which are then taken as nodes for graph. Once a graph is generated, we do a simple search of nodes of graph in aggregated graph. A Yes/No answer checks if all the nodes and edges in the question graph is present in the aggregated graph or not. A contextual questions finds object - edge subgraph in aggregated graph and returns a list of all nodes which are connected to object - edge subgraph. A temporal query finds the first and second question graph in aggregated graph and then checks their time stamps. If time\_stamp\_1 is less than time\_stamp\_2, only then it returns a True.

These node lookups are O(1) or constant time lookups as we are storing the graph as a networkx object (similar to a dict/json representation). Any sequence or subgraph lookups will be O(k+|E|) where k is the length of the sequence in the query and E are the number of edges in the graph.

\section{Experiments}

The experiments we conducted follow the proposed methodology in the same flow. We have built a pipeline which takes a video as an input and outputs a scene graphs for each segmented line.

The pipeline includes all the core tasks:

\subsection{Frame extraction} - Done via \textit{ffmpeg} library which works with video mp4 data.

\subsection{Dense Captioning} - We ran the github implementation of JCJohnson for dense captioning. However this implementation proved to be a bottleneck for real time performance. The implementation involved training over 1.2 GB of pretrained weights. Additionally it took 15 minutes to generate captions for a single image on our local systems. To overcome this we are using an API for dense captioning implemented by DeepAI which significantly reduces overhead time. 

\subsection{Scene Graph Generation} - The scene graphs generation was executed with the help of Stanford CoreNLP library. We also created an executable .jar file for this module which can be called by an outside code across platforms and languages, which helped us in creating an executable pipeline for the methodology. 



\subsection{Scene Graph Similarity and Key Frame Extraction}
As explained in the proposed method, we now analyze and report the results for Scene Graph similarity across frames. We use two similarity measures as mentioned earlier - Spectral similarity and MCS similarity. The first video we analyze for this step is an instructional video (cooking video) and the second is a video of a house party/music video.

\begin{figure}[h!]
\centering
\includegraphics[width = 1\linewidth]{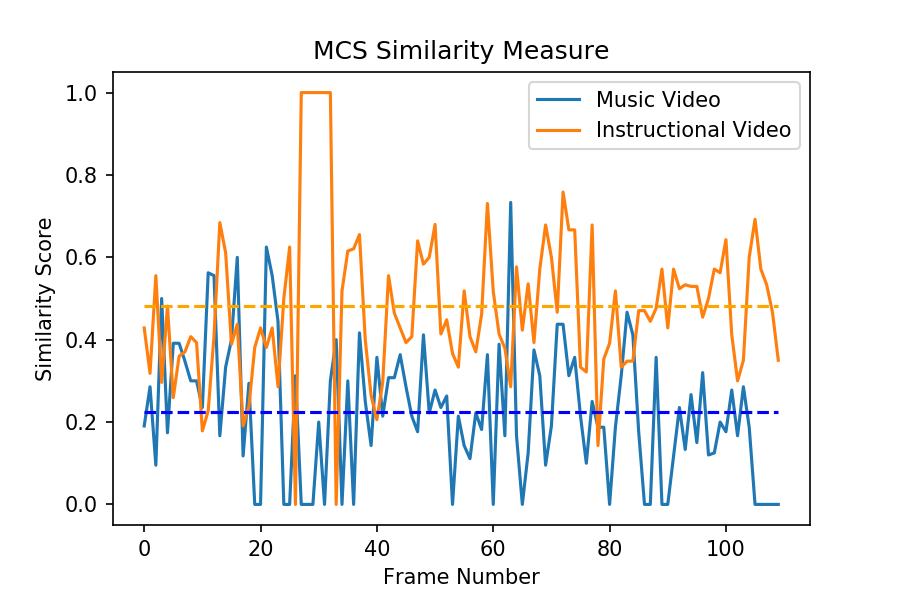}
\caption{Scene Graph similarity - MCS}
\label{fig:simmcs}
\end{figure}

\begin{figure}[h]
\centering
\includegraphics[width = 1\linewidth]{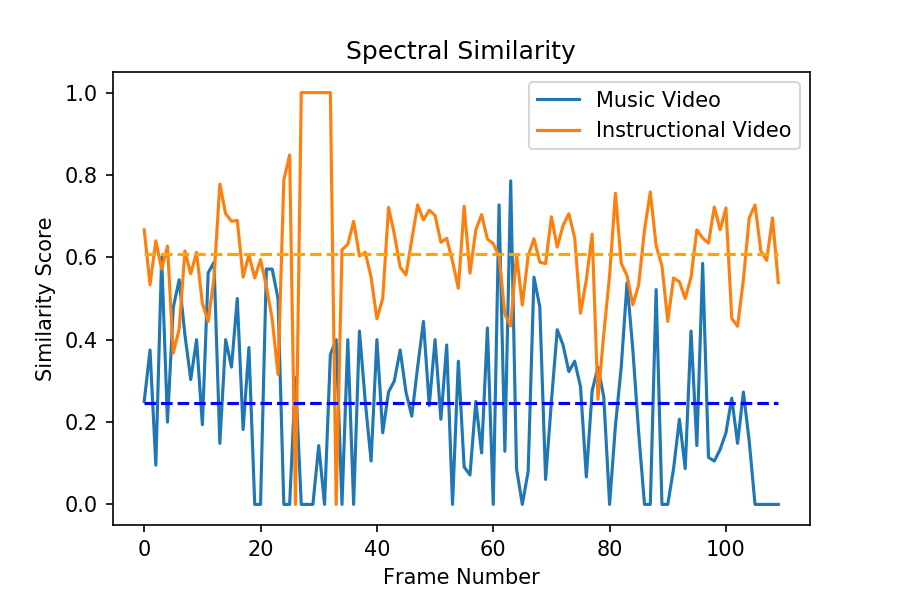}
\caption{Scene Graph similarity - Spectral}
\label{fig:simspec}
\end{figure}

\begin{figure}[h]
\centering
\includegraphics[width = 1\linewidth]{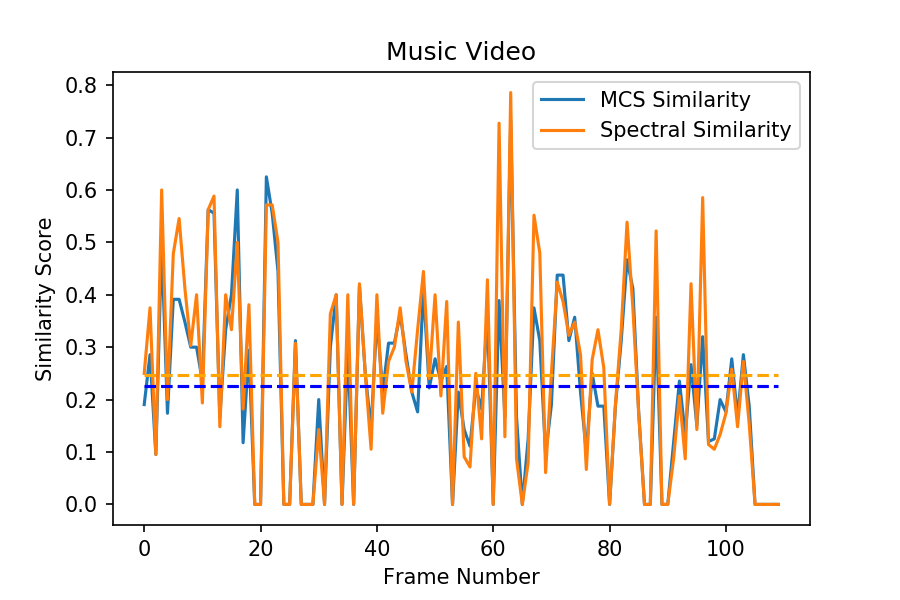}
\caption{Scene Graph Similarity - Music Video}
\label{fig:simmusic}
\end{figure}

\begin{figure}[h]
\centering
\includegraphics[width = 1\linewidth]{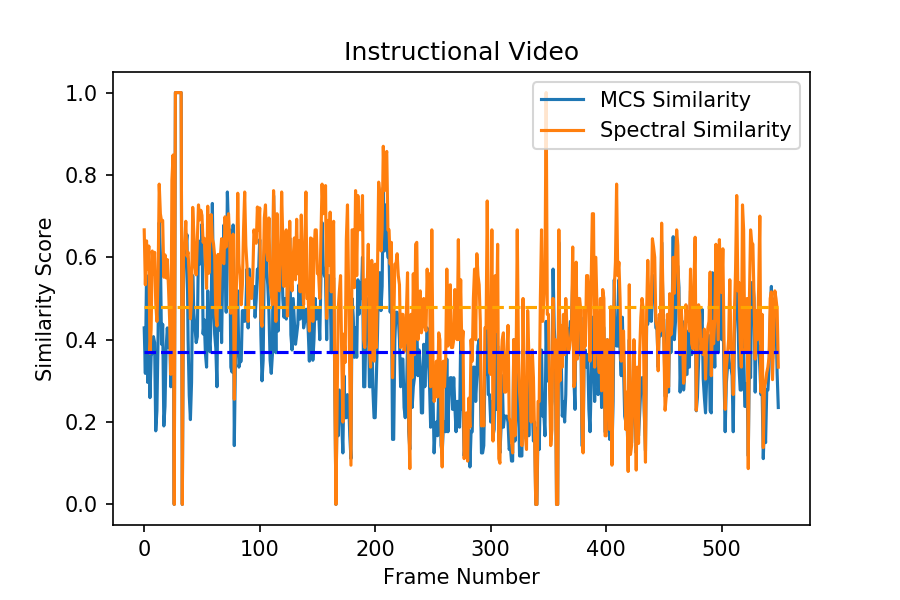}
\caption{Scene Graph Similarity - Instructional Video}
\label{fig:siminstr}
\end{figure}

We can make a few observations from the figures (\ref{fig:simmcs} to \ref{fig:siminstr}). As we can see from both figure \ref{fig:simmcs} and figure \ref{fig:simspec}, the music video on an average tends to have lower similarity across frames. On the other hand, the instructional video has an higher average similarity across frames. This is expected as the music video is varying a lot in context as opposed to the relatively stable context of the cooking video. Another point to be noted is that for a given video, while the actual average similarity may vary, the general trend of similarity across frames is preserved in both similarity measures. Figures \ref{fig:simmusic} and \ref{fig:siminstr} illustrate this point. (Note: figure \ref{fig:siminstr} has been generated for 600 frames as opposed to the previous plots but as part of experiments but the observations still hold).

Now that we have established the similarity scores, we can extract the required number of key-frames. If we want `k' key frames, we extract the `k' largest drops in similarity across the frames. Alternatively we could extract the key-frames based on a fixed similarity threshold (say 0.3 or 0.5).

\begin{figure}[h]
\centering
\includegraphics[width = 1\linewidth]{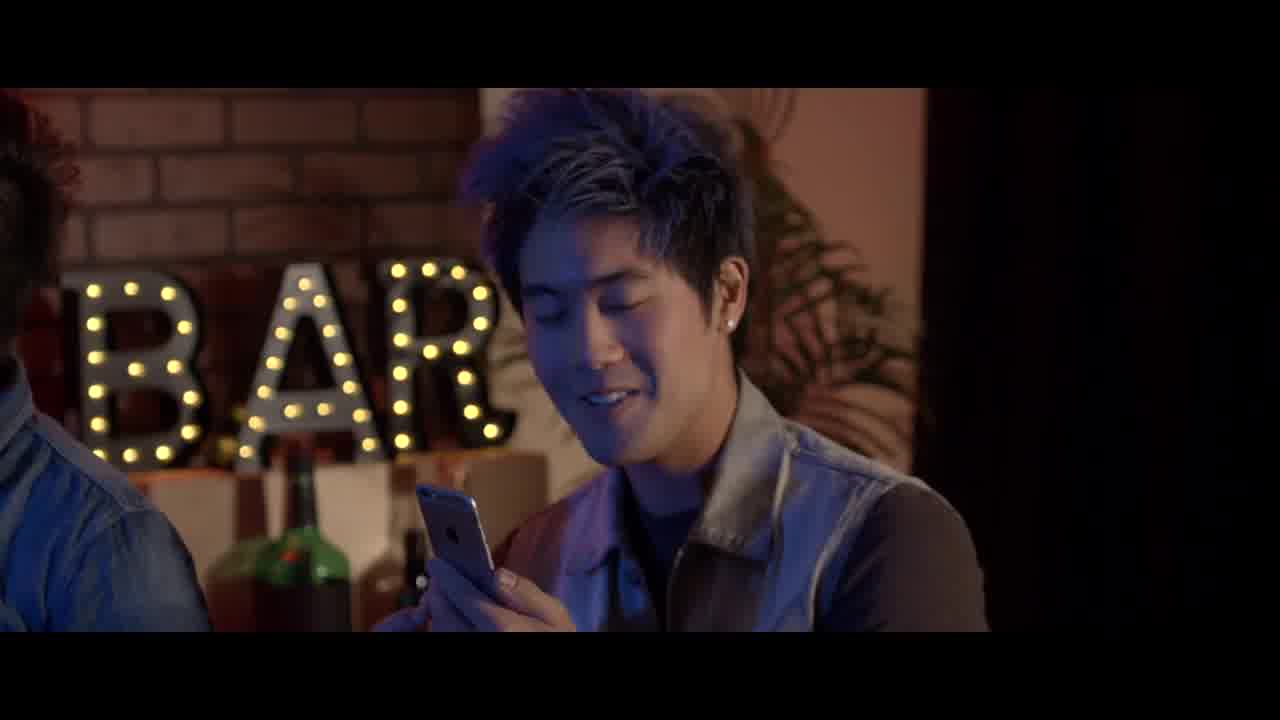}
\caption{Frame 60 - Music Video}
\label{fig:f60music}
\end{figure}

\begin{figure}[h]
\centering
\includegraphics[width = 1\linewidth]{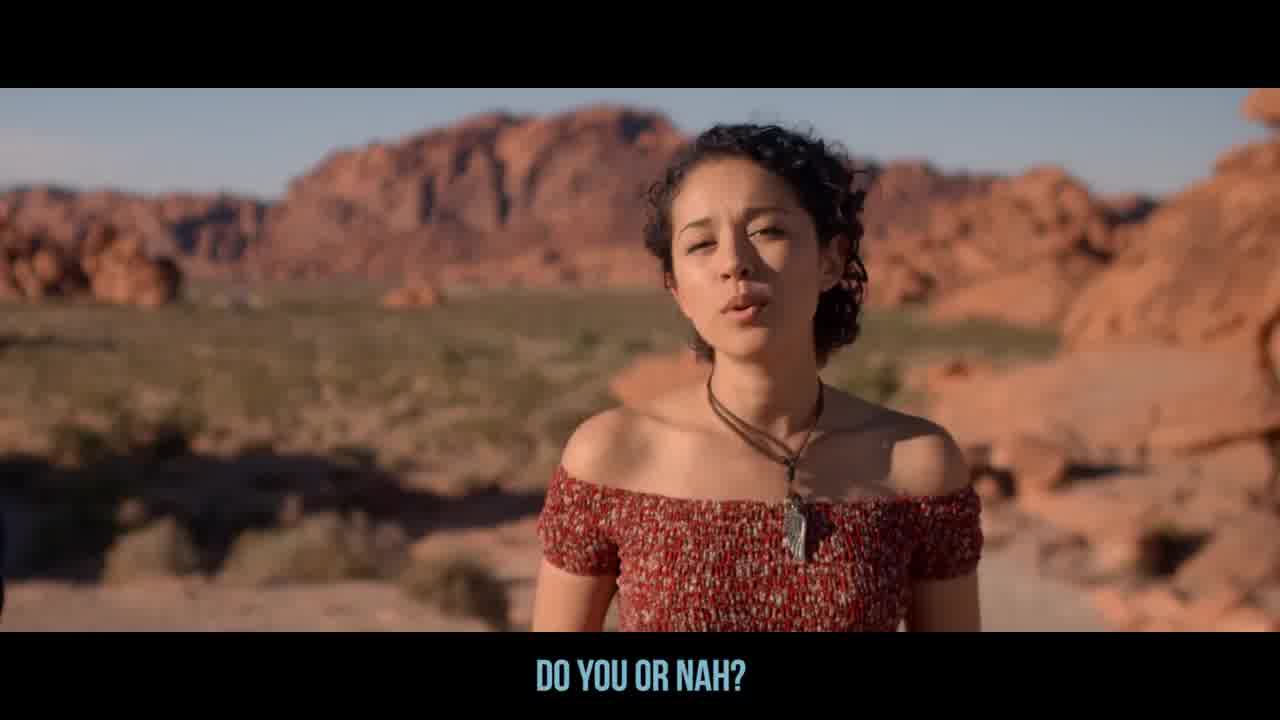}
\caption{Frame 61 - Music Video}
\label{fig:f61music}
\end{figure}

\begin{figure}[h]
\centering
\includegraphics[width = 1\linewidth]{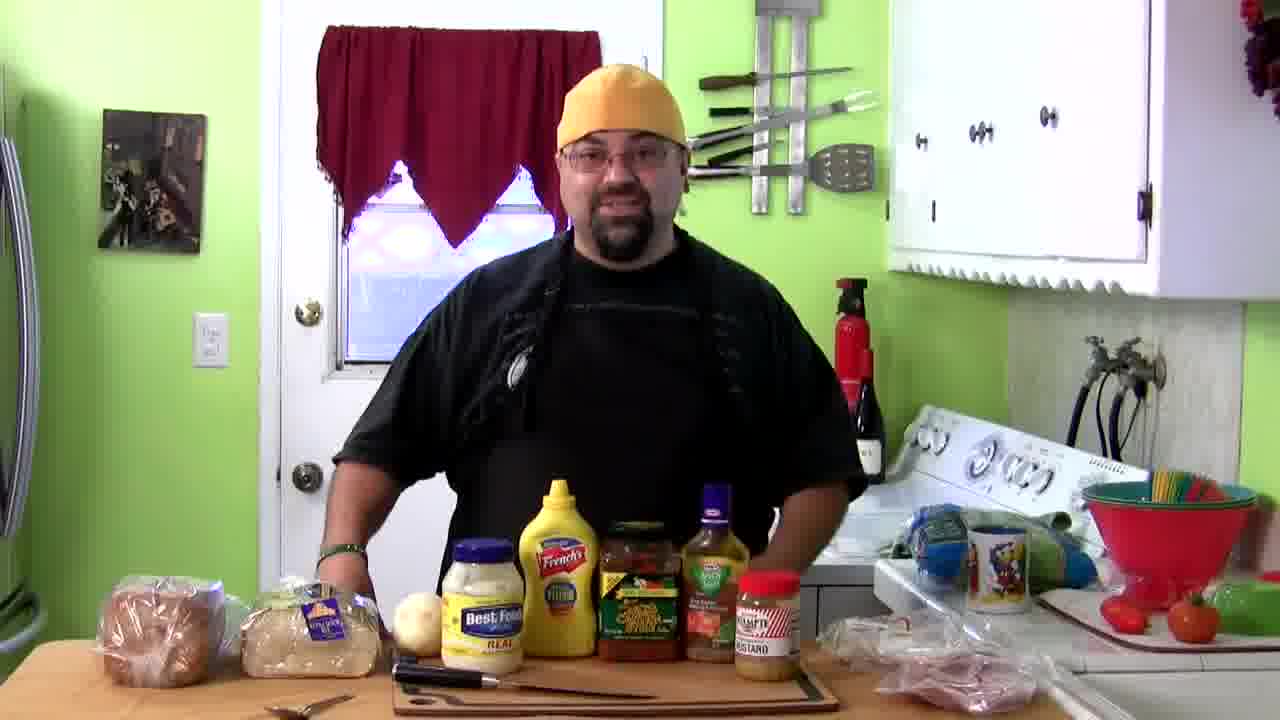}
\caption{Frame 79 - Instructional Video}
\label{fig:f79instr}
\end{figure}

\begin{figure} [h]
\centering
\includegraphics[width = 1\linewidth]{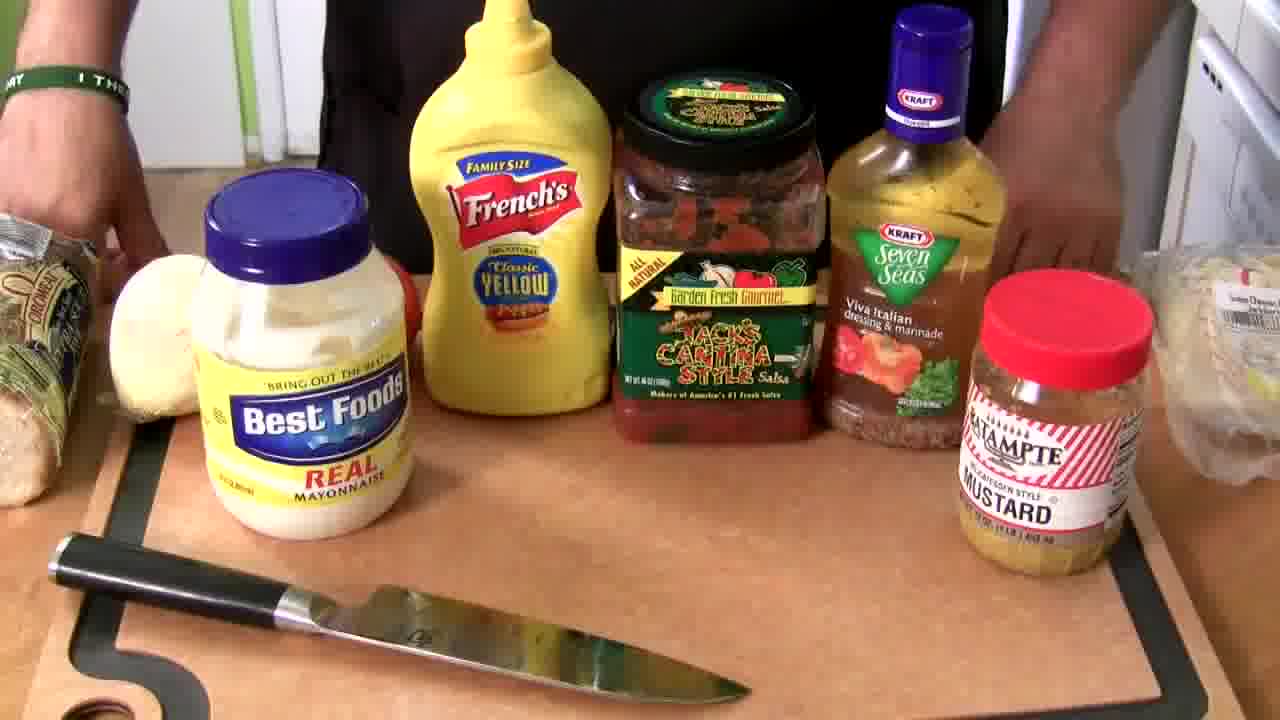}
\caption{Frame 80 - Instructional Video}
\label{fig:f80instr}
\end{figure}

We also show a few scenes where the similarity was low between consecutive frames. For example, in the music video, we see that similarity score is extremely low between frame 60 and 61. We have showed the corresponding frames in figures \ref{fig:f60music} and \ref{fig:f61music}. As we can see, the scene changes drastically and the same is reflected in our similarity score. Similarly, in the instructional video the similarity score is low at between frame 79 and 80. Figures \ref{fig:f79instr} and \ref{fig:f80instr} reinforce this point.

\begin{figure}[h]
\centering
\includegraphics[width = 0.7\linewidth]{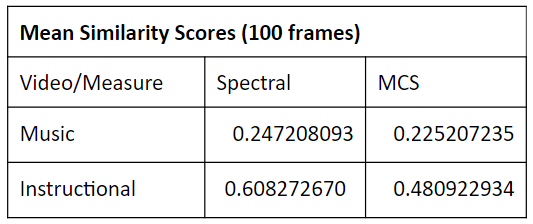}
\caption{Mean Similarity}
\label{fig:meansim}
\end{figure}

\begin{figure}
\centering
\includegraphics[width = 0.7\linewidth]{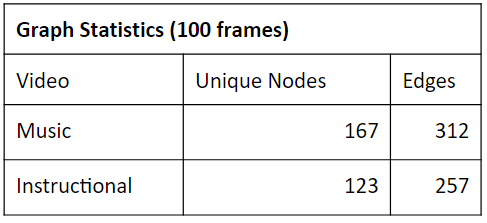}
\caption{Sample Graph Stats}
\label{fig:kfgraphstats}
\end{figure}

Figure \ref{fig:meansim}, \ref{fig:kfgraphstats}, show the similarity score and number od nodes and edges.

\subsection{Scene Graph Aggregation for Query Handling}
After the above step is performed and we have extracted the key frames, we now form an aggregated scene graph from these key frame scene graphs as explained in the node based aggregation section of the method.
Below we show a sample experiment which we performed during prototyping. Here Image1 and Image2 are two key-frames (Scene 1 and Scene 2) extracted as per the above logic.

\subsubsection*{Scene 1}

\begin{figure}[h]
\centering
\includegraphics[width = 1\linewidth]{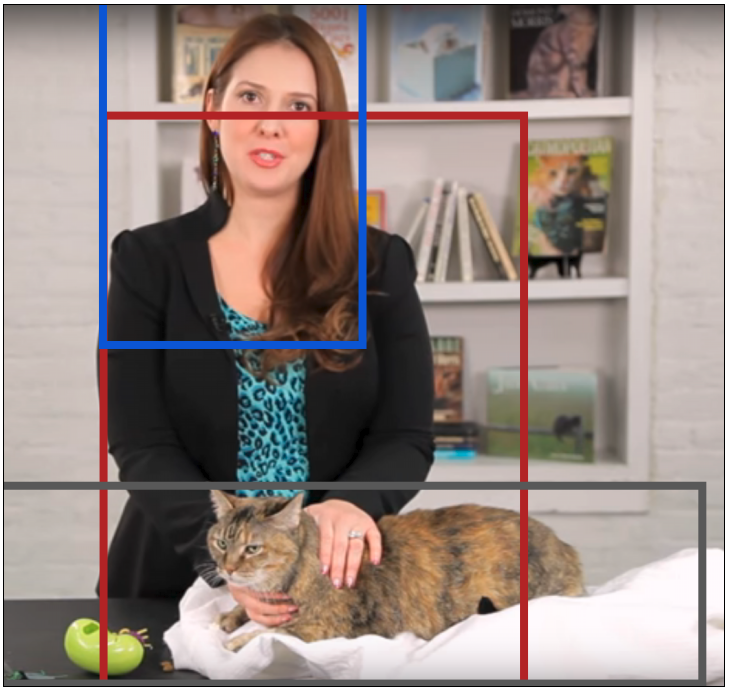}
\caption{Dense Captioning output for Image 1}
\label{fig:agg1}
\end{figure}

The dense captions for the above frame are as follows :
``Woman with long hair'',
``Woman playing with cat''.
``Brown cat sitting on a bench''

Each sentence gives us a graph as shown below and we aggregate them to form the scene graph for the current frame. Note here that the 2 woman nodes were merged into the same node based on the bounding box information.

\begin{figure}
\centering
\includegraphics[width = 1\linewidth]{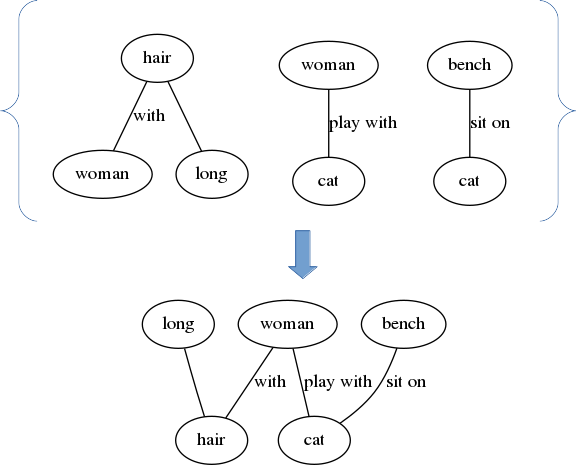}
\caption{Scene graph of Image1}
\label{fig:agg2}
\end{figure}

\begin{figure}
\centering
\includegraphics[width = 1\linewidth]{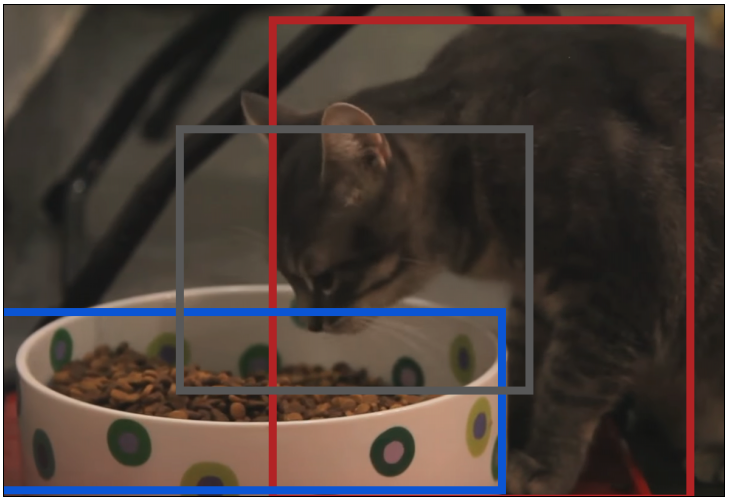}
\caption{Dense Captioning output for Image 2}
\label{fig:agg3}
\end{figure}

\begin{figure}
\centering
\includegraphics[width = 1\linewidth]{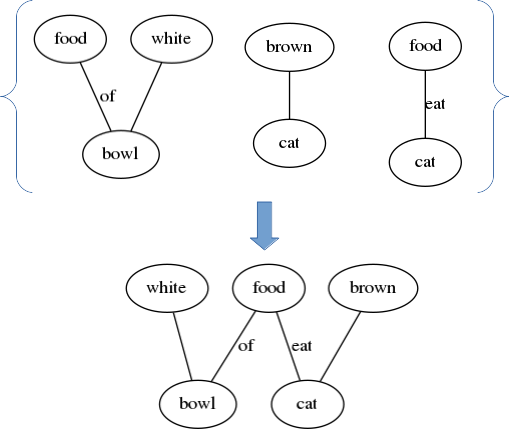}
\caption{Scene graph of Image2}
\label{fig:agg4}
\end{figure}

\begin{figure}
\centering
\includegraphics[width = 1\linewidth]{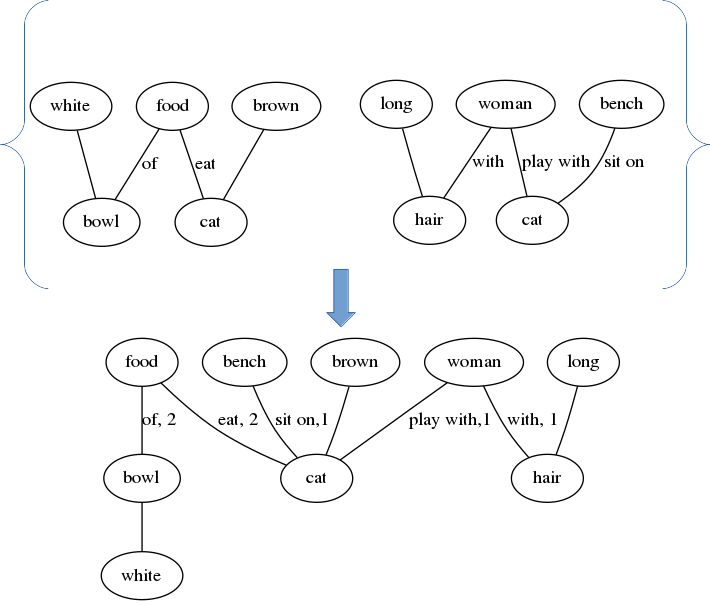}
\caption{Aggregated graph of Image1 and Image2}
\label{fig:agg5}
\end{figure}

Figures \ref{fig:agg1} through \ref{fig:agg4} illustrate this point for both Image1 and Image2. Figure \ref{fig:agg5} illustrates how the two scene graphs were aggregated. As we can see, cat is a common node to both graphs and its attributes and edges are combined.

\subsection{Question Answering}

We now query the aggregated graph and compare with human annotation for accuracy measurement. For the purpose of this project we consider any answer incorrect if it does not match the human annotated answer. Most answers are single worded so we do not need any complex metrics. It is a one-zero metric. Based on this logic, we annotated 4 videos of varying content and report the results below in figure \ref{fig:qastats}.
Note here in the figure that questions titled `What?' and `Where?' are contextual questions while questions titled `When?' are temporal questions.
\begin{figure}[h]
\centering
\includegraphics[width = 0.7\linewidth]{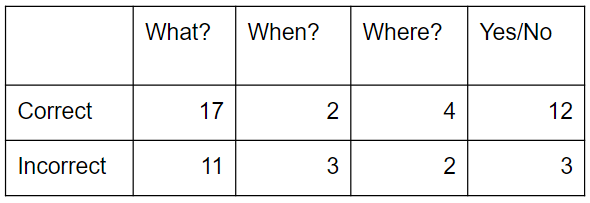}
\caption{Question Answer Stats}
\label{fig:qastats}
\end{figure}

\begin{figure}[h]
\centering
\includegraphics[width = 1\linewidth]{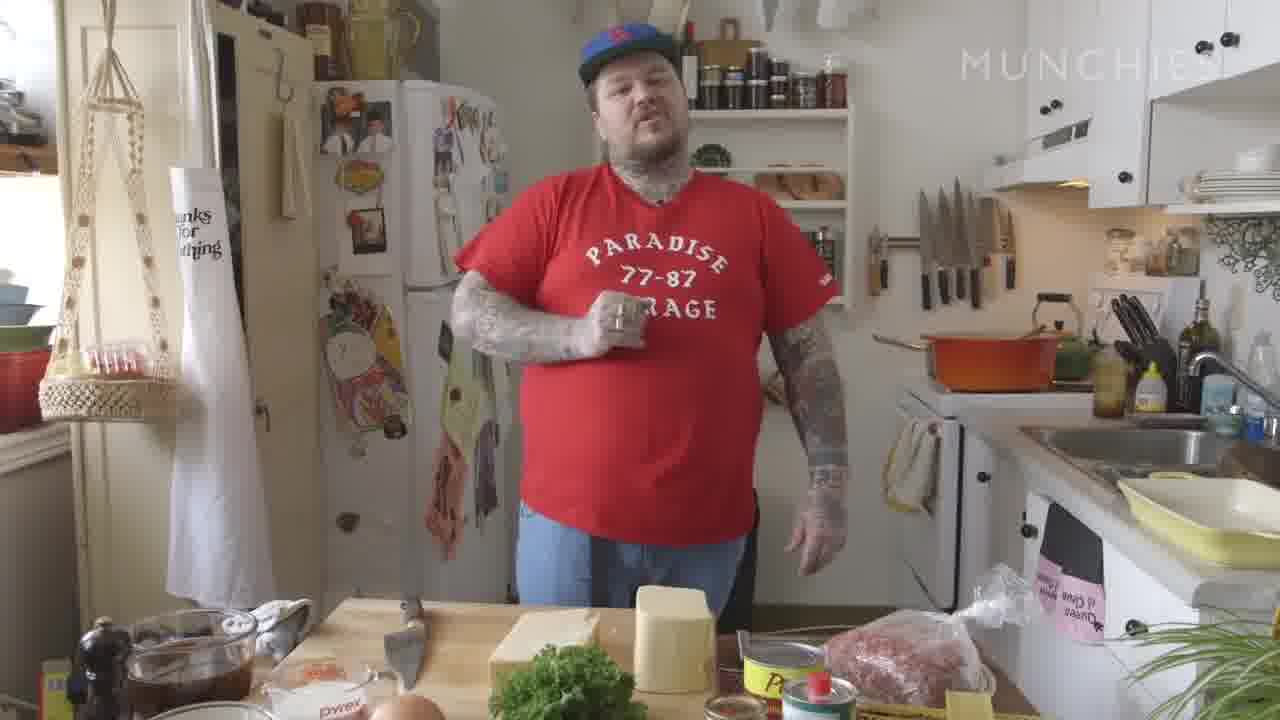}
\caption{Correctly Answered Question - Example 1}
\label{fig:correct1}
\end{figure}

\begin{figure}[h]
\centering
\includegraphics[width = 1\linewidth]{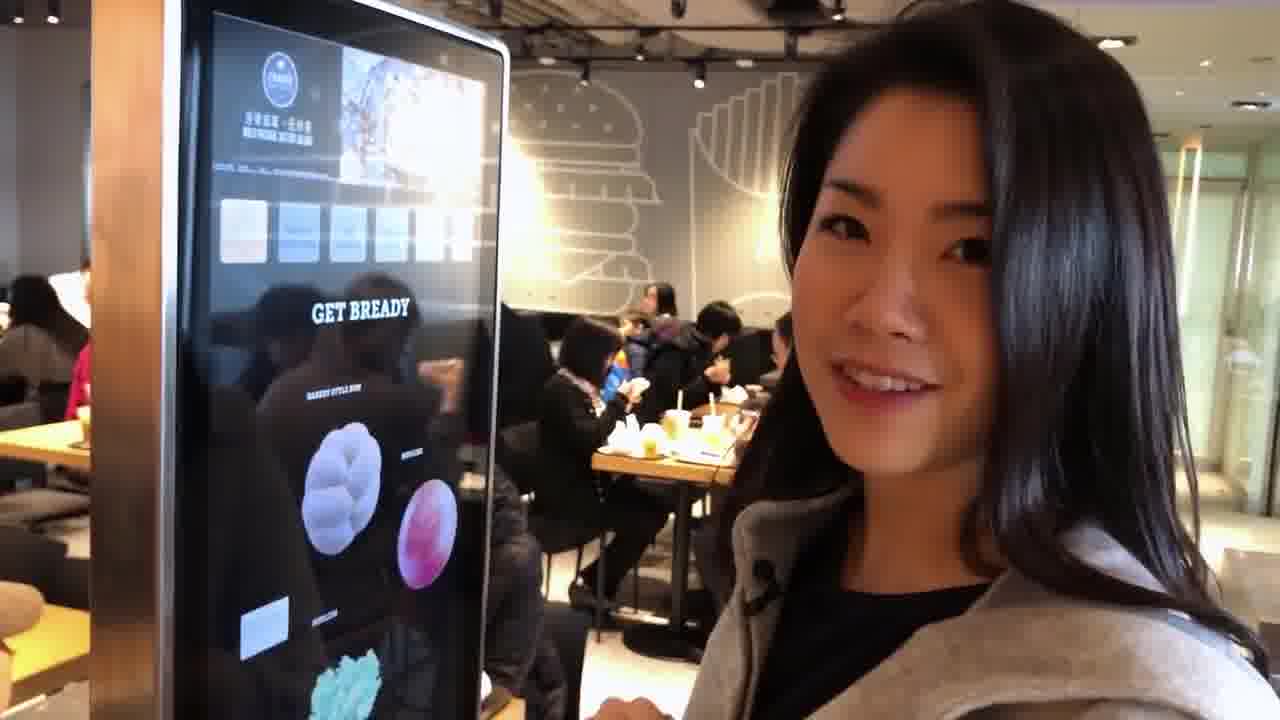}
\caption{Correctly Answered Question - Example 2}
\label{fig:correct2}
\end{figure}

We also report a few examples where the questions were answered correctly along with the actual frame in the video. 
Images for below examples are shown in figures \ref{fig:correct1} and \ref{fig:correct2})

\noindent Q: Where is the man?
\begin{verbatim}A: [['kitchen']] \end{verbatim} 

\noindent Q: What is the woman wearing?
\begin{verbatim}A: [['shirt', 'jacket']] \end{verbatim}

\begin{figure}[h]
\centering
\includegraphics[width = 1\linewidth]{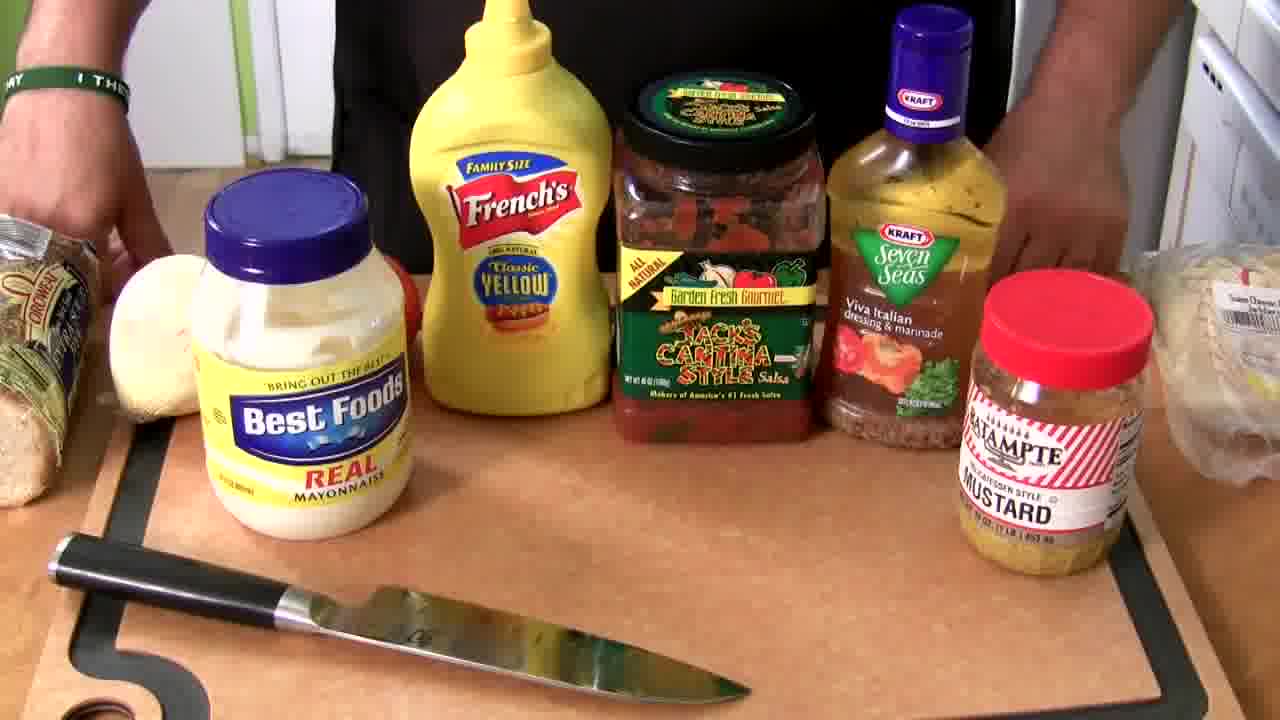}
\caption{Incorrectly Answered Question - Example 1}
\label{fig:incorrect1}
\end{figure}

Among the questions that were unanswered or answered incorrectly, we find that the captioning for that image was incorrect resulting in incorrect answers. Figure \ref{fig:incorrect1} shows the frame for the below incorrect answer.

\noindent Q: What is on the table?
\begin{verbatim}A: [['bottle','jar','beer','water','food']] \end{verbatim}

\subsection{Video Similarity}

\begin{figure}[h]
\centering
\includegraphics[width = 0.7\linewidth]{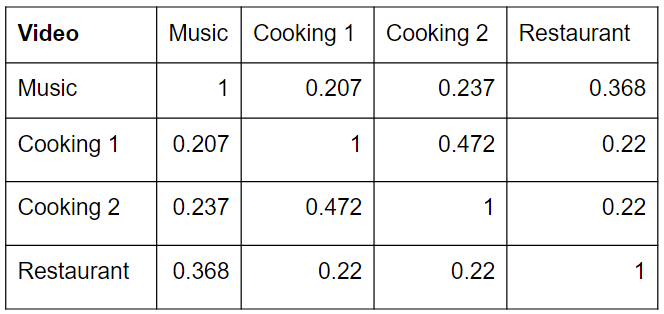}
\caption{Video Similarity scores using Spectral similarity}
\label{fig:videosimfrob}
\end{figure}

\begin{figure}[h]
\centering
\includegraphics[width = 0.7\linewidth]{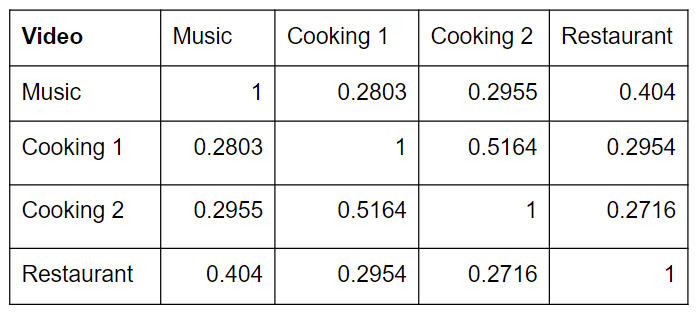}
\caption{Video Similarity scores using MCS similarity}
\label{fig:videosimmcs}
\end{figure}

We now aggregate scene graphs in a given video as per the `bag of nodes' method mentioned earlier. Though such a representation may not retain contextual/temporal relationships, we can use this as a crude measure for video similarity. We use both the Spectral similarity measure and the MCS similarity measure and find the scores as reported in figures \ref{fig:videosimfrob} and \ref{fig:videosimmcs}. 

Video1 -> Music video;
Video2 -> Cooking video 1;
Video3 -> Cooking video 2;
Video4 -> Restaurant video;

As we can see from figures, the similarity scores correspond to what a human would annotate as `similar' videos - for example Cooking Video 1 and Cooking Video 2 have the highest pairwise similarity score. We can see that if we have a video database and calculate pairwise video similarity scores, we can retrieve the closest video to a given video query. 

\begin{figure}
\centering
\includegraphics[width = 1\linewidth]{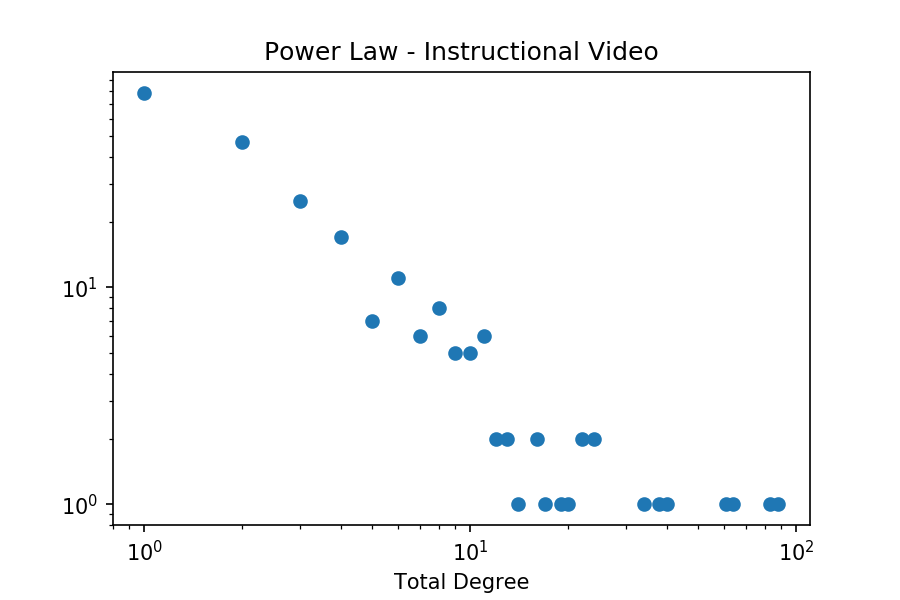}
\caption{Power Law}
\label{fig:power}
\end{figure}

Using the `bag of nodes' graph aggregation method, we also observe an approximate power law distribution for total degree. We visualize this for a sample video as seen in figure \ref{fig:power}.

\section{Future Directions}

\begin{enumerate}
\item
The biggest bottleneck of the whole pipeline is the dense cap algorithm. This algorithm captures the relationships and objects from  images. A better trained dense cap network will produce much better captions which will produce much richer node attributes and catch much more relationships between nodes.
\item
Optimized graph building ways - We append each scene graph with a timestamp to existing scene graph. Each node is checked if it already exists in the graph by using similarity measures based on attributes. This is a brute force way of node similarity comparison. If time permits, we will look into a more optimized way of aggregate graph generation.  

\item
Graph similarity measures - For a single image, we get multiple scene graphs based on segmentation. Similarity for a single frame is based on node similarity based on node attribute comparison and bounding box information. Across frames, node similarity is based on node attribute comparison only as objects can move across various frames. It would be an interesting problem to track people across multiple frames.

\item
Currently we are handling different queries on different graphical representations. While non-temporal queries are handled in an aggregated graph, temporal queries are answered by querying the individual scene graphs for all key frames. In the future, we can formulate a single graphical representation capable of handling all queries.

\item
We are currently considering only the visual data of a given scene. Future work can build up on this by also incorporating audio data to improve the contextual awareness of the question-answering model.

\item
We have briefly touched upon video similarity in this paper. Future work in this direction could build upon this crude similarity metric for unsupervised clustering of videos and video retrieval based on an input video without using any video meta-data.

\end{enumerate}

\section{Related Work}
Existing literature describes a lot on building Contextual Question Answering on images. Related work described below explored ideas on conversion of scenes to graphs, which captures the context present in the scene.

Firstly, creating an efficient representation of input data which
retains the contextual relationships between them. While [6] explored
methods to generate semantically precise scene graphs and
improve upon the state of the art, [5] achieved the same goal using
pixel values of the image, which is a novel technique. In parallel, [8]
proposed building a knowledge base which preserved the relation
between objects as well as focused on solving inherent scalability
issues of previous related works.

Secondly, papers [1] and [3] focused on improving multi object
annotation of images. These papers described models which learn
representation of images in natural language domain from the data
instead of relying upon hard-coded templates. Although images
were not explicitly converted into scene graphs, the annotations
generated can be seen as analogous to scene graphs. Additionally
these works also space-localized the objects within the image.

Thirdly, we explore papers which focused on techniques in efficient
image retrieval. The papers reviewed covered multiple solutions
to the task. [2] retrieved a relevant image given a scene
graph improving upon traditional methods. On the other hand [1]
handled text queries to retrieve images and solved a new problem
of retrieving localized portions within the image corresponding to
the query.

Finally, we look at synthetic image generation given a text description.
Current state of the art this field in this new field is
explored in papers [4] [7].

The common theme across these papers is to associate textual
and visual representations. These papers provide novel mechanisms
or improve state of the art to transform information from one
domain to another. Another recurring theme is in the mechanisms
itself which rely heavily on deep learning and machine learning
Existing shortcomings in traditional methods served as motivation
to these papers. The challenges undertaken by these authors
were across multiple areas such as availability of abundant labeled
data, relating work from traditionally orthogonal fields of NLP and
CV, performance issues associated with deep neural networks, extending
classical techniques and algorithms to contextually aware
algorithms, modeling the complex interactions between objects in
a scene and generalizing to objects not present in the current data

\section{Conclusion}
We implemented an efficient model for answering contextual questions based on videos. The model was built using scene graphs which allows to encode relationships between subjects like man, dog etc. in a graph. The data included videos of varying context as well lengths on which queries were asked in order to test the efficiency of the model. We also introduced a novel way to extract key frames which rely on scene graphs and capture higher contextual changes. The results demonstrate the effectiveness of this method to extract key-frames based on the theory proposed as well as their saliency of being described as key-frames. Finally, we looked at video similarity using  scene graph representation and we propose a future application where similar videos can be clustered and recommended based on content similarity as opposed to tags which are the current way of recommending videos. Building a recommendation system based our proposed approach will allow for more relevant recommendations. 

One major take-away from the project is the proof of concept of the effectiveness of a graph approach of building a query engine to answer contextual questions based on specific videos, which till now has been approached from a deep learning only point of view.  

\section{Work Division}
The algorithm development was done by brainstorming session in which everyone contributed equally. The report was done as a joint effort where each person contributed to the parts they primarily worked on. Though the below areas indicate the area one most contributed to, each person was involved in some capacity in all areas of the project.
\begin{enumerate}
\item
Akash Ganesan: Q\&A engine, core NLP work, visualization, pipeline, graph and search combinators.
\item
Shubham Dash: Pipeline, dense captioning, graph linking, Q\&A model validation.
\item
Karthik Muthuraman: Key frame extraction using graph similarity and visualization, graph linking, Q\&A data collection.
\item
Divyansh Pal: Dense captioning, graph linking, Q\&A data collection.
\end{enumerate}


\nocite{*}

\bibliographystyle{ACM-Reference-Format}
\bibliography{sample-bibliography}

\end{document}